\documentclass[10pt,twocolumn,letterpaper]{article}

\usepackage[pagenumbers]{cvpr}

\usepackage{adjustbox}
\usepackage{graphicx}
\usepackage{amsmath}
\usepackage{amssymb}
\usepackage{booktabs}
\usepackage[table]{xcolor}
\usepackage{bm}
\usepackage{pifont}
\usepackage{bbding}
\usepackage{multirow}
\usepackage{xspace}
\usepackage{tabularx}
\usepackage{siunitx}
\newcommand{\cmark}{\ding{51}}
\newcommand{\xmark}{\ding{55}}
\definecolor{Gray}{gray}{0.90}
\definecolor{citecolor}{RGB}{34,139,34}

\usepackage[pagebackref,breaklinks,colorlinks]{hyperref}

\usepackage[capitalize]{cleveref}
\crefname{section}{Sec.}{Secs.}
\Crefname{section}{Section}{Sections}
\Crefname{table}{Table}{Tables}
\crefname{table}{Tab.}{Tabs.}

\newcommand{\methodName}{SCALE\xspace}

\newcommand{\mcm}{\text{MCM}\xspace}
\newcommand{\mcmExtended}{Masked Clip Modeling\xspace}

\usepackage{stackengine}
\usepackage{scalerel}
\def\mcirc{\scalebox{.9}{$\bigcirc$}}
\def\nmediumcircle{\kern-.1pt\mcirc\kern-8.9pt\mcirc}

\def\ncincircle{\mathbin{%
  \stackengine{.2pt}{\nmediumcircle}{\scalebox{.96}{$\text{\texttt c}$}}{O}{c}{F}{T}{L}}}

\def\cincircle{\mathbin{%
  \ThisStyle{\scalebox{.87}{$\SavedStyle\scalerel*{\ncincircle}{b}$}}}}

\begin{document}

\title{Spatio-Temporal Crop Aggregation for Video Representation Learning}

\author{Sepehr Sameni\\
Computer Vision Group\\
University of Bern\\
{\tt\small sepehr.sameni@unibe.ch}
\and
Simon Jenni\\
Adobe Research\\
\\
{\tt\small jenni@adobe.com}
\and
Paolo Favaro\\
Computer Vision Group\\
University of Bern\\
{\tt\small paolo.favaro@unibe.ch}
}
\maketitle

\begin{abstract}
   We propose Spatio-temporal Crop Aggregation for video representation LEarning (\methodName), a novel method that enjoys high scalability at both training and inference time. Our model builds long-range video features by learning from sets of video clip-level features extracted with a pre-trained backbone. To train the model, we propose a self-supervised objective consisting of masked clip feature prediction. We apply sparsity to both the input, by extracting a random set of video clips, and to the loss function, by only reconstructing the sparse inputs. Moreover, we use dimensionality reduction by working in the latent space of a pre-trained backbone applied to single video clips. These techniques make our method not only extremely efficient to train but also highly effective in transfer learning. We demonstrate that our video representation yields state-of-the-art performance with linear, non-linear, and $k$-NN probing on common action classification and video understanding datasets.
\end{abstract}

\section{Introduction}
\label{sec:intro}

Videos provide rich and detailed information about objects and their activities. Their analysis is, however, made challenging not only by the difference in the information provided across space and time but also by the high dimensionality of the data~\cite{recasens2021broaden,huang2021ascnet,zhang2012slow}.
While computational resources are expected to scale over time, so is the demand for higher data resolution (both in space and time) and also the need for processing data with even more dimensions, such as videos of volumetric data~\cite{choy20194d}. Therefore, it is of paramount importance to explore methods that drastically reduce the computational requirements to process videos.
Moreover, the annotation of videos is an extremely costly and time-consuming burden that makes the use of models pre-trained in a self-supervised manner essential~\cite{ranasinghe2022self}.

Self-supervised learning (SSL) is a very popular technique to reduce the need for annotation because it can build useful representations from unlabeled data through an artificial goal, also called pseudo- or pretext-task. These representations can either be evaluated through K-nearest neighbor or linear probing~\cite{zhou2021ibot,zhou2022mugs} or through fine-tuning (i.e., as the initialization parameters of the trained model)~\cite{he2022masked,tong2022videomae,liu2022exploring} on a downstream task, where only a small labeled dataset is available. More remarkably, SSL pre-trained models can outperform models that were pre-trained on an annotated dataset~\cite{tomasev2022pushing,Azizi2021BigSM}.

In the case of videos, SSL methods for video representation learning present fundamental scalability challenges~\cite{tong2022videomae,recasens2021broaden,huang2021ascnet,wu2022memvit}. 
A first major challenge is that training models from scratch for any new pseudo-task is neither feasible, sustainable, nor scalable.
A more viable setting is one where data, such as videos, is pre-processed only once via some pre-trained general-purpose model (e.g., trained via self-supervised learning \cite{feichtenhofer2021large,grill2020bootstrap,he2020momentum,ranasinghe2022self,caron2021emerging,lorre2020temporal,oord2018representation}), and the (compressed) representation is stored and used later for other training purposes or retrieval.
A fundamental question is whether we can improve the performance of video representations by training a model on top of pre-computed features. A second challenge is that even a single round of training can be quite demanding.
However, a lot of the video content is redundant and sparsity could be used to reduce the computational cost~\cite{tong2022videomae}.

To make the learning of video representations highly scalable, we propose a method that works on four fronts:
\begin{enumerate}
    \item \textbf{Input Sparsity:} 
    Sparsity in the input to the model~\cite{tong2022videomae, he2022masked, assran2022masked,girdhar2022omnimae,Akbari2021VATTTF} is an effective way to drastically reduce the computational load and memory requirements, while taking advantage of the information redundancy in images and videos~\cite{feichtenhofer2022masked}. Inspired by prior work, we extract a sparse set of clips, instead of image patches or video tubelets, from a video. Each clip is then fed separately to a neural network to obtain a video clip representation.
    
    \item \textbf{Output Sparsity:} 
    Another way to reduce the computational cost is to use a sparse reconstruction output instead of a dense one \cite{zhou2021ibot,tan2021vimpac,Wang2022BEVTBP,Bao2022BEiTBP}. This is not just a reduction of the number of terms in the loss function, but also in the number of actual outputs of the model, and thus also a reduction of the computations needed to obtain them. This is in contrast to MAE-based SSL methods for vision, where the proposed pseudo-tasks are based on the reconstruction of the whole input~\cite{he2022masked,tong2022videomae,girdhar2022omnimae,feichtenhofer2022masked}, and even if the loss uses a subset of the tokens (the masked ones) for the loss calculation, the remaining tokens are still part of the decoder's output and computation graph.
    
    \item \textbf{Dimensionality Reduction:} Inspired by prior work~\cite{zhou2021ibot,oord2018representation,Dong2022BootstrappedMA}, instead of directly processing the raw input data, we work in the latent space. This allows us to further reduce the dimensionality of both input and output data.

    \item \textbf{Use of a Pre-trained Backbone:} To reduce training time and further speed up the processing per iteration, we exploit SSL pre-trained models. These pre-trained backbones already learn very strong short-term spatiotemporal features, and our approach is a way to extract longer-term features (see LVU results~\ref{table:lvu}) by building a video representation on top of a set of pre-trained features (one for each video clip).
\end{enumerate}

To integrate all these components, we propose a novel SSL method that we call Spatio-temporal Crop Aggregation for video representation LEarning (\methodName).
Given an input video, \methodName extracts a random set of clips and produces an embedding for each clip through a pre-trained backbone, which is kept frozen.
These initial embeddings are then augmented in two ways: 1) each one is refined into a more discriminative feature, and 2) the set of all embeddings is summarized in a global feature.
These global features can learn long-term correlations in the whole video by aggregating the short-term information in each clip embedding.
The combination of the initial embeddings with their refinement and the global feature is then used in an ensemble for applications on new downstream tasks.

To train \methodName, we introduce two novel pseudo-tasks, which aim to improve the discriminability of the embeddings of each video clip as well as obtain a global representation of the video. 
One task, which we call \mcmExtended (\mcm), is the reconstruction of a video clip embedding as in masked autoencoders \cite{tong2022videomae}. Masked embeddings are combined with positional encodings so that the model can (spatio-temporally) relate the missing input embeddings to the other available embeddings (similarly to BERT in Natural Language Processing~\cite{devlin2018bert}). A second task is to train the model to output a global feature token (which we refer to as CLS, just for consistency with previous works~\cite{caron2021emerging,devlin2018bert,dosovitskiy2020image,Touvron2021TrainingDI,chen2021empirical}) for a set of clips via contrastive learning, so that the global feature is invariant to the chosen set of clips from the same video, but can discriminate summary features of clips from other videos. Both tasks are trained via contrastive losses.

\methodName consistently improves upon the pretrained features and the significance of the improvements becomes even more evident when one considers the computational cost that other methods require. For instance, \methodName would achieve a performance improvement of 0.3\% over the best VideoMAE~\cite{tong2022videomae} (1600 epochs checkpoint -- fine-tuning) and of 0.6\% in the case of the best $\rho$BYOL~\cite{feichtenhofer2021large} (800 epochs checkpoint -- linear probe) on Kinetics400 with 64 V100 GPUs in about \textbf{5 minutes}. In contrast, as a reference and with the same computational resources, VideoMAE requires about \textbf{27.7 hours} to improve its performance of 0.5\% through fine-tuning from its 800 to 1600 epochs checkpoint, and $\rho$BYOL needs at least \textbf{48 hours} to improve of 0.4\% its linear probing performance from its 200 to 400 epochs checkpoint.

To summarize our contributions, we propose \methodName, a novel and highly scalable video representation method that
\begin{itemize}
    \item is trained via novel pseudo-tasks on sets of video clips (in contrast to existing methods that work only with pairs of clips \cite{ranasinghe2022self,feichtenhofer2021large,recasens2021broaden} at a time);
    \item results in video feature representations with a significant performance improvement in $k$-NN (retrieval), linear, and non-linear probing across a wide range of datasets for action classification (UCF~\cite{soomro2012ucf101}, HMDB~\cite{kuehne2011hmdb}, SSv2~\cite{goyal2017something}, Kinetics400~\cite{kay2017kinetics}) and long-form video understanding (LVU~\cite{Wu2021TowardsLV});
    \item achieves consistent transfer learning performance improvement across different SotA pre-trained backbones (in terms of architectures, scale, and pre-training tasks). For example our non-linear probed model even outperforms the fully fine-tuned SVT~\cite{ranasinghe2022self} on HMDB~\cite{kuehne2011hmdb}.
\end{itemize}

\section{Related Work}
\label{sec:background}
Our approach relates to many prior works on (self-supervised) representation learning. 
In particular, \methodName relates to SSL approaches on videos, methods that rely on multiple views of the data, and predictive methods, where part of the data is predicted from another part.

\noindent\textbf{Self-Supervised Video Representation Learning.} 
Early SSL approaches on video were based on pseudo tasks, \eg, the recognition of transformations of video frame sequences \cite{misra2016shuffle,benaim2020speednet,wei2018learning}. 
More recently, popular methods developed on images have been successfully translated to video, \eg, contrastive methods \cite{feichtenhofer2021large,grill2020bootstrap,he2020momentum}, clustering-based methods \cite{ranasinghe2022self,caron2021emerging}, or predictive approaches \cite{lorre2020temporal,oord2018representation}. 
Often, these approaches are tailored to video by including additional learning signals, \eg, by combining contrastive methods with temporal constraints \cite{dave2022tclr} and pretext tasks \cite{jenni2021time}, or by incorporating audio \cite{morgado2021audio}, or optical flow \cite{han2020self}.
These approaches typically learn representations with limited temporal extent, which can serve as backbones for our approach. 

\noindent \textbf{Learning from Multiple Views.} 
Many methods have explored using multiple views (\eg, generated through space-time cropping) to represent and learn from videos.
For example, many SSL approaches rely on multiple views of the data, \eg, in contrastive formulations \cite{feichtenhofer2021large,ranasinghe2022self,Patrick2021SpaceTimeC}, or predictive learning \cite{recasens2021broaden}, where invariance to views is the goal.
Other approaches aim to learn from the relation between two views, \eg, by predicting overlap \cite{zhang2022contrastive}, the relative distance \cite{sun2021composable}, or cross-view feature prediction \cite{yuan2022contextualized,tao2022siamese} and reconstruction \cite{nash2022transframer}.
Besides exploiting multiple views for SSL, some works also propose general multi-view video models, \eg, by capturing and fusing features at different spatiotemporal resolutions \cite{yan2022multiview}, by aggregating information over longer time-spans \cite{wu2019long,sener2020temporal,wu2022memvit,wang2022long}, or by selecting important frames \cite{gowda2021smart}.
These approaches, however, are learned end-to-end. In contrast, we propose a more scalable approach by learning a global video representation of pre-trained features extracted over multiple crops using self-supervision. 

\noindent \textbf{Predictive Learning.} 
One of our proposed SSL objectives is a prediction of space-time crop features given other video crop features.
This approach is similar to masked token prediction as in BERT~\cite{devlin2018bert} and relates to several other methods in the literature.
Masked input reconstruction methods have recently become popular on images \cite{he2022masked} and successfully translated to video \cite{tong2022videomae,feichtenhofer2022masked}.
Other approaches formulate masked prediction tasks in the learned feature space \cite{zhou2021ibot,tao2022siamese,Dong2022BootstrappedMA} or some fixed latent space \cite{tan2021vimpac}.
Another line of work considers directional predictions (\eg, into the future) often formulated via contrastive predictive coding \cite{oord2018representation} applied to video \cite{lorre2020temporal, liu2022contrastive, suris2021learning, wu2020learning, girdhar2021anticipative}. 
These masked prediction tasks are typically formulated on a fixed grid (\eg, at the level of patches or frames). In contrast, our formulation is continuous, predicting features of arbitrarily sampled space-time crops.

\section{Scalable Video Representation Learning}
\label{sec:method}
To describe \methodName, we first define some basic notation and functions, including a general contrastive loss notation that we use for all training losses.

\subsection{Notation} 
We use lower-case letters (e.g., $z$) to denote generic vectors and capital letters (e.g., $Z$) to denote their sets. The expression $a\cincircle b$ denotes the concatenation of $a$ and $b$. We also skip writing the parameters of networks (usually denoted by $\theta$) if their presence and role are clear from the context. Throughout the description of the method, we refer to the training of neural networks, and, therefore, at each iteration of the training, we sample a minibatch of videos. All the equations in the next sections are written for a single video in the minibatch. Although we do not explicitly indicate it, all the contrastive losses also use the other videos in the minibatch as negatives.

\subsection{Contrastive Loss}
InfoNCE is a powerful method for representation learning~\cite{oord2018representation} that can be used to maximize the mutual information between two variables. Because we use this loss between different variables throughout our method, we introduce here a unified notation.
Let the paired sets $A$ and $B$ have $N$ elements each, $A=\{a^{i}\}_{i=1}^{N}$ and $B=\{b^{i}\}_{i=1}^{N}$, where $a^{i}$ are vectors of dimension $d_A$ and $b^{i}$ are vectors of dimension $d_B$. We also introduce two Multi Layer Perceptrons (MLP), parameterized with $\theta_{A}$ and $\theta_{B}$, to project these vectors onto a common space of dimension $d$.
After feeding the elements $a^i$ and $b^i$ to the MLPs and normalizing them, we obtain
\begin{equation}
     \Tilde{a}^{i} = \frac{\mbox{MLP}_{\theta_{A}}(a^{i})}{\lVert \mbox{MLP}_{\theta_{A}}(a^{i}) \rVert}\quad\text{and}\quad \Tilde{b}^{i} = \frac{\mbox{MLP}_{\theta_{B}}(b^{i})}{\lVert \mbox{MLP}_{\theta_{B}}(b^{i}) \rVert},
\end{equation}
where $\|\cdot\|$ denotes the $L_2$ norm.
We define the per-element loss based on the relative similarity of $\Tilde{a}^{i}$ and $\Tilde{b}^{i}$, and by using a temperature $\tau$
\begin{equation}
    \Tilde{\ell^{i}}(A, B, \theta_{A}, \theta_{B}) = 
    -\log\frac{\exp\left(\frac{\Tilde{a}^{i} \cdot \Tilde{b}^{i}}{\tau}\right)}{
    \displaystyle
    \sum_{j=1}^{N}
    \textstyle \exp\left(\frac{\Tilde{a}^{i} \cdot \Tilde{b}^{j}}{\tau}\right)}.
\end{equation}
We then make the loss symmetric~\cite{Radford2021LearningTV} by
\begin{equation}
    \ell^{i}(A, B, \theta_{A}, \theta_{B}) = 
    \Tilde{\ell^{i}}(A, B) + \Tilde{\ell^{i}}(B, A)
\end{equation}

Finally we define the contrastive loss ${\cal L}_\text{cntr}(A, B, \theta_{A},\theta_B)$ as the mean of $\ell^{i}$
\begin{equation}
     {\cal L}_\text{cntr}(A, B, \theta_{A}, \theta_{B}) = \frac{1}{N}  \sum_{i=1}^{N}{\ell^{i}}(A, B, \theta_{A}, \theta_{B}).
\end{equation}
As mentioned earlier on, for simplicity, in the rest of the paper we will not indicate the parameters of the MLPs, and simply write ${\cal L}_\text{cntr}(A, B)$ or $\ell^i(A,B)$.

\begin{figure*}[t]
    \centering
    \includegraphics[width=\textwidth,trim=0 0 7cm 0,clip]{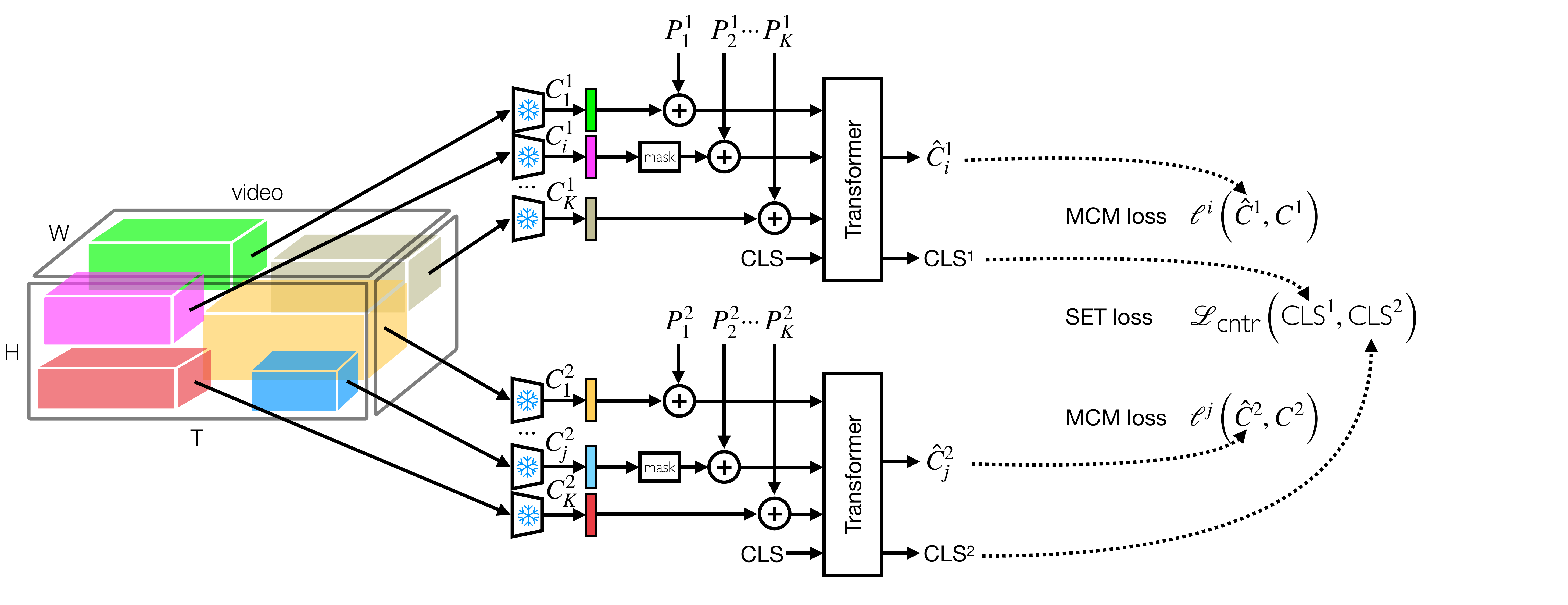} 
    \caption{Video Representation Learning with \methodName. For each video, \methodName extracts two sets of video clips $V_1^1,\dots,V_K^1$ and $V_1^2,\dots,V_K^2$. Each video clip is processed separately through a frozen backbone $\text{E}_{\mbox{\scriptsize \ding{100}}}$ and results in encoded video clips $C_1^1,\dots,C_K^1$ and $C_1^2,\dots,C_K^2$. Then, a random set of encodings in each set is masked and reconstructed at the output of the predictor network (a transformer) ($\ell^i$). The predictor network is also fed a class token $\text{CLS}$. The corresponding output token encodes a summary $\text{CLS}^m$ of the $m$-th set of video clips. The objective for these summary tokens is to be similar only when encoding video clips from the same video ($\mathcal{L}_\text{SET}$).}
    \label{fig:scale}
\end{figure*}

\subsection{Training \methodName}
In our method, we integrate $4$ principles to drastically reduce the computational complexity of learning a video representation: Input sparsity, output sparsity, dimensionality reduction, and use of a pre-trained backbone. 
Moreover, we introduce two pseudo-tasks to train the model. One task is based on the (contrastive) reconstruction of a masked video clip given some context video clips. The second task is to build a global representation that is (contrastively) invariant to the set of input sampled video clips. The overall training scheme of \methodName is illustrated in Figure~\ref{fig:scale}.

\noindent\textbf{Input sparsity.} 
As a first step, rather than processing a whole video, we collect a sparse set of short video clips from the same video. Given a video $V\in\mathbb{R}^{H\times W \times T \times 3}$, where $H$, $W$ and $T$ are the height, width, and duration (in frames) of the video, we sample $2K$ clips. We divide the clips into two sets randomly. Each clip in the first set $V^1_{i}\in \mathbb{R}^{H_{i}^1\times W_{i}^1 \times T_{i}^1}$, with $i=1,\dots,K$, is obtained at the spatio-temporal location $X_i^1,Y_i^1,Q_i^1$ with different data augmentations and dimensions $H_{i}^1$, $W_{i}^1$ and $T_{i}^1$. Similarly, we denote the second set of clips $V^2_{i}$, for $i=1,\dots,K$.
We also normalize their coordinates relative to the video dimensions and embed them onto a feature vector $P_i^j$ by feeding them to a learnable MLP. We denote these embeddings 
\begin{align}
\textstyle
    P^{j}_{i}=\mbox{MLP}\left(\left[\frac{X^j_{i}}{H},\frac{Y^j_{i}}{W},\frac{Q^j_{i}}{T},\frac{X^j_i+H^j_{i}}{H},\frac{Y^j_i+W^j_{i}}{W},\frac{Q^j_i+T^j_{i}}{T}\right]^\top\right),
\end{align}
where $j=1,2$ and $i=1,\dots,K$.

\noindent\textbf{Dimensionality reduction and pre-trained backbone.} 
To reduce the dimensionality of each video clip, we feed them independently to a frozen encoder $\text{E}_{\mbox{\scriptsize \ding{100}}}$, to obtain the encodings $C^{j}_{i}=\text{E}_{\mbox{\scriptsize \ding{100}}}(V^{j}_{i})$, where $j=1,2$ and $i=1,\dots,K$.
Our framework is encoder agnostic and thus can work with encoders obtained through different training schemes (supervised, contrastive, or autoencoder). In addition to reducing the dimensionality, we make the training even more scalable by using pre-trained and frozen encoders. Note, however, that if performance is the main goal, it is possible to also train a sparse backbone end to end with multiple clips. Thanks to token dropping, one can drop up to 95\% of the tokens~\cite{girdhar2022omnimae} and still build a good representation.

\noindent\textbf{Output sparsity.}
As a self-supervised signal for our video representation learning, we use a (contrastive) reconstruction loss. To reduce the computational cost, instead of predicting the features for the whole video~\cite{Wang2022BEVTBP,girdhar2022omnimae,tong2022videomae,feichtenhofer2022masked} (asymmetric decoding), we only reconstruct a \emph{sparse} set of masked clips.
Our reconstruction objective is based on the observation that video signals carry a lot of redundancy. Hence, we introduce a model, the \emph{predictor network}, to predict masked video clip embeddings given the other video clip embeddings (the context). 
We follow the general approach of BERT~\cite{devlin2018bert} but implement the predictor network as a masked autoencoder, where the reconstruction is based on a contrastive loss. 
The loss is applied only to a sparse set $M^1 \subset \{1,\ldots,K\}$ of masked video clips. These clips are replaced by a learned $\text{MSK}$ token. All embeddings $C^1_i$, including the masked ones, are added to their corresponding position encoding $P^1_i$ and are then fed to the predictor network. We also include an additional learnable $\text{CLS}$ token as input for the predictor network, which will be used for tasks with multiple video clips.
We denote the outputs of the predictor network as $\hat C^1_i$ for the tokens corresponding to $C^1_i$, and as $\text{CLS}^1$ for the token corresponding to $\text{CLS}$.
Similarly, we feed as inputs separately from the previous set all the video clips $C^2_i$ with their corresponding positional embeddings $P^2_i$, and the same $\text{CLS}$ token, and obtain $\hat C^2_i$ and $\text{CLS}^2$ respectively (see Figure~\ref{fig:scale} for a visual depiction of these processing steps).

\noindent\textbf{Contrastive reconstruction.} Modeling all the details of a masked clip, even in the latent space and even given the redundancy in videos, is a demanding task. Rather than increasing the capacity of our model, since we are aiming for scalability, we keep our predictor network a (relatively) shallow network and use contrastive learning~\cite{oord2018representation}. With contrastive learning, the predicted representation of the masked tokens should only be closer to the original unmasked clip representation (after an MLP projection) than from all other clips from the same video and the rest of the minibatch. 
Note that the rest of the clips in the same video act as hard negatives in contrastive learning~\cite{Robinson2021ContrastiveLW}. Also, since we are using a frozen backbone, we can afford to use large minibatch sizes, which is known to be beneficial for contrastive learning~\cite{chen2021empirical}.
We call this contrastive reconstruction loss the \mcmExtended (\mcm) loss
\begin{equation}\label{eq:mcm_loss}
     \mathcal{L}_{\mcm} = \sum_{i\in M^1}\ell^i(\hat C^1, C^1) + \sum_{j\in M^2}\ell^j(\hat C^2, C^2).
\end{equation}

\noindent\textbf{Multiple video clips loss.} 
The predictor network outputs features for each video clip that are highly discriminative. This task is similar to that of a masked autoencoder \cite{he2022masked} and gives you an enhanced per-clip representation. For many video tasks, we need a global representation for the whole video; for that, we introduce an additional pseudo-task that captures a more global representation of a set of video clips.
Our task takes inspiration from contrastive learning methods used in SSL, which yield representations that perform well with linear probing~\cite{chen2020simple}. The loss aims to make the $\text{CLS}^1$ and $\text{CLS}^2$ tokens returned from the predictor network more similar (recall that these two tokens are obtained from two separate groups of video clips extracted from the same video) than to other class tokens from other videos within the minibatch
\begin{equation}\label{eq:set_loss}
     \mathcal{L}_\text{SET} = {\cal L}_\text{cntr}\left(\text{CLS}^1, \text{CLS}^2\right).
\end{equation}
In addition to our contrastive loss (InfoNCE), one can use clustering~\cite{caron2021emerging} or regression~\cite{grill2020bootstrap} losses. We chose InfoNCE for simplicity and for better compatibility between the losses. As the overall loss, we used the sum of both loss terms (without any weights)
\begin{equation}\label{eq:full_loss}
     \mathcal{L} =\mathcal{L}_{\mcm} + \mathcal{L}_\text{SET},
\end{equation}

\section{Experiments}
\label{sec:experiments}
We evaluate \methodName on several commonly used action classification datasets for video representation learning. 
As our performance metric, we primarily use linear probing and non-linear probing~\cite{Hjelm2019LearningDR}. 
For the smaller datasets, we also use $k$-NN classifiers (which are training-free) and demonstrate that the proposed method improves upon both unsupervised and supervised backbones.

\subsection{Experimental Setup and Protocols}
\label{subsec:exp_setup}

\noindent\textbf{Computational Efficiency:}
In Table~\ref{table:gpu}, we show estimates of the maximum batch sizes and the training throughput for different methods trained with the same computational and memory resources. As can be seen, \methodName is orders of magnitude more efficient than other SotA methods.

\begin{table}[t]
	\centering\small
    \resizebox{\linewidth}{!}{
    \begin{tabularx}{0.45\textwidth}{lcccc|c}
		\toprule 
		Batch Size 
        & 8 & 19 & 57 & 2048 & Samples/s\\
		\midrule
		MoCo$^\text{V3}$ 
        & 23.47 & OOM & OOM & OOM & 8.66\\
		VMAE 
        & 11.42 & 22.80 & OOM & OOM & 33.54\\
		MoCo$^\text{V3}_\text{Sparse}$ 
        & 5.49 & 9.27 & 23.62 & OOM & 28.83\\
		\methodName 
        & 1.21 & 1.23 & 1.58 & 19.52 & 9224.65\\
		\bottomrule
	\end{tabularx}
    }
    \caption{\textbf{Computational Resources.} GPU RAM usage (in GB) and max training speed (using one 3090 GPU) of ViT$_\text{B}$ with different batch sizes and different SSL tasks for videos. MoCo$^\text{V3}_\text{Sparse}$ is similar to MSN~\cite{assran2022masked}. OOM indicates Out Of Memory error.}
    \label{table:gpu}
\end{table}

\noindent\textbf{Datasets:} Following prior work~\cite{ranasinghe2022self, feichtenhofer2021large,recasens2021broaden} we use Kinetics-400~\cite{kay2017kinetics}, UCF-101~\cite{soomro2012ucf101} (split 1), HMDB-51~\cite{kuehne2011hmdb} (split 1), and Something-Something v2 (SSv2)~\cite{goyal2017something} to train and evaluate our models. We also use the LVU benchmark~\cite{Wu2021TowardsLV} to showcase our long-form video understanding capabilities. 
Note that almost 35\% of LVU videos are not available to download from YouTube anymore; thus, our results are not directly comparable with prior methods.

\noindent\textbf{Pretrained backbones:} We use the pretrained checkpoints of $\rho$BYOL~\cite{feichtenhofer2021large}, SVT~\cite{ranasinghe2022self}, and three variants of VideoMAE~\cite{tong2022videomae} (base(B), large(L), and fine-tuned base(FT)). We choose $\rho$BYOL for their excellent linear performance, SVT for the usage of ViT~\cite{dosovitskiy2020image}, and VMAE for showing 1) the applicability of our proposed method to MAE models, 2) the scalability of our method to larger models, and 3) possibility of using supervisedly fine-tuned models as our backbone. All the models are self-supervisedly pretrained on Kinetics-400, except the fine-tuned VMAE base that was also supervisedly finetuned on Kinetics-400. We also used a backbone pretrained and fine-tuned on SSv2 (VMAE$^\text{B}_\text{SSv2}$) for the SSv2 experiment to show the universality of \methodName with respect to the pretraining dataset.

\newcommand{\pacc}[1]{{\bf \fontsize{7.5}{42}\selectfont \color{citecolor!80}~(#1)}}
\begin{table*}[t]
    \centering
    \resizebox{\textwidth}{!}{
    \begin{tabular}{@{}cclllllll@{}}
        {} & {} & Relation & Speak & Scene & Director & Genre & Writer & Year\\
        \toprule
        \textcolor{gray!80}{SlowFast+NL} & \cite{Feichtenhofer2019SlowFastNF, wang2018non} & \textcolor{gray!80}{52.40} & \textcolor{gray!80}{35.80} & \textcolor{gray!80}{54.70} & \textcolor{gray!80}{44.90} & \textcolor{gray!80}{53.00} & \textcolor{gray!80}{36.30} & \textcolor{gray!80}{52.50}\\
        \textcolor{gray!80}{ViS4mer} & \cite{Islam2022LongMC} & \textcolor{gray!80}{57.14} & \textcolor{gray!80}{40.79} & \textcolor{gray!80}{67.44} & \textcolor{gray!80}{62.61} & \textcolor{gray!80}{54.71} & \textcolor{gray!80}{48.80} & \textcolor{gray!80}{44.75}\\
        \midrule
        \multirow{5}{*}{SVT} & Linear & 64.70 & 35.77 & 62.33 & 37.27 & 54.35 & 52.12 & 28.46\\
        {} & \methodName$_\text{linear}$ & 73.52\pacc{$+$8.82} & 40.65\pacc{$+$4.88} & 68.83\pacc{$+$6.50} & 46.36\pacc{$+$9.09} & 57.94\pacc{$+$3.59} & 56.38\pacc{$+$4.26} & \bf{39.23}\pacc{$+10.77$}\\
        \cline{2-9}
        {} & MLP & 67.64 & 39.02 & 66.23 & 45.45 & 56.92 & 57.44 & 36.15\\
        {} & Transformer & 70.58 & 40.65 & 68.83 & 47.27 & 57.17 & 58.51 & 36.92\\
        {} & \methodName$_\text{ft}$ & \bf{76.47}\pacc{$+$5.89} & \bf{42.27}\pacc{$+$1.62} & \bf{74.02}\pacc{$+$5.19} & \bf{49.09}\pacc{$+$1.82} & \bf{58.97}\pacc{$+$1.80} & \bf{62.76}\pacc{$+$4.25} & \bf{39.23}\pacc{$+$2.31}\\
        \hline
        \multirow{5}{*}{$\rho$BYOL} & Linear & 52.94 & 38.21 & 53.24 & 36.36 & 51.79 & 57.44 & 33.84\\
        {} & \methodName$_\text{linear}$ & \bf{67.64}\pacc{$+$14.70} & 43.08\pacc{$+$4.87} & 66.23\pacc{$+$12.99} & 44.54\pacc{$+$8.18} & 53.33\pacc{$+$1.54} & 60.63\pacc{$+$3.19} & 40.00\pacc{$+$6.16}\\
        \cline{2-9}
        {} & MLP & 62.35 & 41.46 & 62.42 & 47.27 & 52.56 & 59.57 & 40.00\\
        {} & Transformer & 65.09 & 44.71 & 66.23 & 50.90 & 53.07 & 61.70 & 43.84\\
        {} & \methodName$_\text{ft}$ & \bf{67.64}\pacc{$+$2.55} & \bf{45.52}\pacc{$+$0.81} & \bf{71.42}\pacc{$+$5.19} & \bf{51.81}\pacc{$+$0.91} & \bf{55.72}\pacc{$+$2.65} & \bf{65.95}\pacc{$+$4.25} & \bf{46.92}\pacc{$+$3.08}\\
        \bottomrule
    \end{tabular}
    }
    \caption{\textbf{Long-form Video Understanding Results.} Linear and non-linear probing accuracies on LVU~\cite{Wu2021TowardsLV} classification tasks. 
    We see a significant performance boost for \methodName compared to the baselines, indicating \methodName's ability to capture long-form video features.
    Because the dataset is not fully available to download, the methods in the first two rows cannot be directly compared with \methodName and are thus greyed out.}
    \label{table:lvu}
\end{table*}

\noindent\textbf{Self-supervised Training:} For training data, we extract 16 clips of 16 frames from each video per dataset and save their feature encodings to disk. We use PySlowFast's common data augmentations for that~\cite{fan2020pyslowfast}. For evaluation, we follow the 5x3 scheme~\cite{Feichtenhofer2019SlowFastNF} (uniformly sampling 5 clips
from a video along its temporal axis and then taking 3 spatial crops) and extract 15 clips from each video (except for SSv2, where we extract 2x3 clips~\cite{Wang2019TemporalSN}).
As the architecture for the predictor network, we use an encoder-only Transformer~\cite{vaswani2017attention} and a three-layer MLP (without batch normalization~\cite{ioffe2015batch}) for the contrastive heads. Unless stated otherwise, we train our models for 500 epochs (for example, training with VMAE$^\text{B}$ on SSV2 takes 137 minutes with one 3090 GPU) with a batch size of 512 and use all 16 clips.

\noindent\textbf{Evaluation:}
Since our focus is on efficient and scalable video classification, we always freeze the backbones in our evaluation (as in our self-supervised pretraining) and either train a linear classifier~\cite{ranasinghe2022self,feichtenhofer2021large} or fine-tune the predictor network (the transformer) with an additional linear head. 
Therefore, when we refer to fine-tuning (ft), we \emph{only} adapt the non-linear head (\eg, predictor network) but \emph{not the backbone}.
We apply a grid search for the hyper-parameters of the heads covering learning rate, weight decay, batch size, and optimizer type. 
Similar to MAE~\cite{he2022masked}, we found that applying a batch normalization layer~\cite{ioffe2015batch} without affine transformations is beneficial for VideoMAE models. As the linear baseline, we consider the well-established ensembling approach, \ie, we average the softmax predictions of the 15 clips (6 for SSv2) to obtain the final prediction. For models that process multiple clips at once (like ours), we likewise apply a linear softmax head on the concatenation of the individual clip features and the [SET] token before averaging to obtain the final prediction. 
For the smaller datasets, we also use $k$-NN classification, where, similar to DINO~\cite{caron2021emerging}, we always use $k=20$ and work with l2 normalized representations.

\noindent\textbf{Non-linear baselines:} As we are a non-linear model, we consider an MLP on top of the frozen backbone as a non-linear baseline.  As a further baseline
, we consider a Transformer trained on all the clip representations. 
This Transformer uses the exact same architecture as \methodName, and  only differs in the initialization: in the case of \methodName we start from our proposed SSL pre-trained weights instead of random initialization.

\begin{table}[t]
	\centering\small
   \resizebox{\linewidth}{!}{%

	\begin{tabularx}{0.5\textwidth}{@{}l@{\hskip 0.4em}c@{\hskip 0.5em}c@{\hskip 0.5em}c@{\hskip 0.5em}c@{\hskip 0.5em}c@{\hskip 0.5em}c@{}}
		\toprule 
		{} & SVT & $\rho$BYOL & VMAE$^\text{B}$ & VMAE$^\text{L}$ & VMAE$^\text{B}_\text{ft}$ & VMAE$^\text{B}_\text{SSv2}$\\
		\midrule
		Linear & \num{20.30} & \num{25.30} & \num{18.31} & \num{27.94} & \num{28.90} & \num{70.53} \\
		\methodName$_\text{linear}$ & \num{25.26} & \num{27.16} & \num{21.24} & \num{30.18} & \num{33.25} & \underline{\num{70.63}} \\
		\midrule
		MLP & \num{21.43} & \num{26.46} & \num{19.42} & \num{27.96} & \num{29.83} & \num{70.52} \\
		Transformer & \underline{\num{29.24}} & \underline{\num{30.99}} & \underline{\num{24.26}} & \underline{\num{34.39}} & \underline{\num{35.60}} & \num{70.57} \\
		\methodName$_\text{ft}$ & \bf{\num{29.68}} & \bf{\num{31.83}} & \bf{\num{25.25}} & \bf{\num{36.34}} & \bf{\num{37.38}} & \bf{\num{70.69}} \\
  		\bottomrule
	\end{tabularx}
 }
 \caption{\textbf{SSv2 Results.} Linear and non-linear probing accuracies on SSv2~\cite{goyal2017something}. We see that both \methodName$_\text{linear}$ and \methodName$_\text{ft}$ outperform other methods and improve the classification accuracies by a large margin. We also see that \methodName$_\text{ft}$, with its better initialization, always outperforms the Transformer. VMAE$^\text{B}_\text{SSv2}$ was pretrained and fine-tuned on SSv2.}
    \label{table:ssv2}
\end{table}

\subsection{Results}
\label{subsec:results}

\noindent\textbf{LVU:} One of the benefits of \methodName is that it can be used to process long videos, even though the backbones were trained on short videos only. 
To demonstrate the ability to capture long-form video features, we evaluated \methodName on LVU~\cite{Wu2021TowardsLV}, a benchmark that involves seven classification (and two regression) tasks on minute-long videos. Past studies~\cite{Islam2022LongMC, wang2018non} have established that increasing the input's time span enhances accuracy in this challenging dataset. As shown in Table~\ref{table:lvu}, our experiments indicate that \methodName can improve the baseline model's performance by a considerable margin. Moreover, we found that fine-tuning \methodName can lead to further enhancements.

\noindent\textbf{SSv2:} Multiple classes in SSv2 share similar backgrounds and only differ in motion~\cite{Hu2021ContrastAO}, suggesting that high performance on this dataset demonstrates that the model has captured strong motion-related contextual cues~\cite{ranasinghe2022self}. Results in Table~\ref{table:ssv2} show that we outperform the state-of-the-art. 
On this dataset, we see a large performance gap between models that process single clips at a time (Linear and MLP) and the models that work with multiple clips (\methodName and Transformer).
We can see \methodName$_\text{linear}$ is also outperforming the MLP, showing that \methodName is able to capture motion and long-form temporal features of the video. We even improve the supervised model trained on SSv2 (VMAE$^\text{B}_\text{SSv2}$).

\begin{table}[t]
	\centering\small
   \resizebox{\linewidth}{!}{%

    \begin{tabularx}{0.5\textwidth}{@{}lccccc@{}}
		\toprule 
		{} & SVT & $\rho$BYOL & VMAE$^\text{B}$ & VMAE$^\text{L}$ & VMAE$^\text{B}_\text{ft}$\\
		\midrule
		$k\text{-NN}$ & \num{87.20} & \num{85.19} & \num{35.05} & \num{49.14} & \num{96.82} \\
		\methodName$_{k\text{-NN}}$ & \num{89.00} & \num{83.47} & \num{65.63} & \num{76.02} & \num{97.38} \\
		\midrule
		Linear & \num{91.27} & \num{89.55} & \num{66.53} & \num{84.53} & \num{97.91} \\
		\methodName$_\text{linear}$ & \underline{\num{92.65}} & \num{91.43} & \underline{\num{74.46}} & \num{86.78} & \underline{\num{98.14}} \\
		\midrule
		MLP & \num{91.17} & \num{93.60} & \num{71.97} & \underline{\num{87.04}} & \num{98.04} \\
		Transformer & \num{92.20} & \underline{\num{94.34}} & \num{68.22} & \num{86.30} & \num{98.04} \\
		\methodName$_\text{ft}$ & \bf \num{92.94} & \bf \num{95.00} & \bf \num{76.07} & \bf \num{89.92} & \bf \num{98.46} \\
		\midrule
		FT$_\text{reported}$ & \num[round-mode=places,round-precision=1]{93.7} & \num[round-mode=places,round-precision=1]{95.4} & \num[round-mode=places,round-precision=1]{96.1} & - & - \\
		\bottomrule
	\end{tabularx}
 }
 \caption{\textbf{UCF Results.} Linear and non-linear probing accuracies on UCF-101~\cite{soomro2012ucf101}. \methodName$_\text{ft}$ outperforms all the other models and, in the case of $\rho$BYOL, even gets performance close to a fully finetuned model.
 Also, in most cases, \methodName$_\text{linear}$ outperforms the fine-tuned Transformer and achieves state-of-the-art results in linear probing (previous SotA using RGB frames was 92.6~\cite{recasens2021broaden}). We further see a significant accuracy improvement in $k$-NN probing, especially for pre-trained MAE-based models. 
 As a point of reference, the current best fully fine-tuned accuracy (which is not comparable with our setting) is 96.8\%~\cite{wang2022long}. 
 }
    \label{table:ucf}
\end{table}

\begin{table}[t]
	\centering\small
  \resizebox{\linewidth}{!}{%

	\begin{tabularx}{0.5\textwidth}{@{}lccccc@{}}
		\toprule 
		{} & SVT & $\rho$BYOL & VMAE$^\text{B}$ & VMAE$^\text{L}$ & VMAE$^\text{B}_\text{ft}$\\
		\midrule
		$k\text{-NN}$ & \num{51.83} & \num{49.67} & \num{21.96} & \num{29.21} & \num{72.81} \\
		\methodName$_{k\text{-NN}}$ & \num{56.01} & \num{51.56} & \num{37.18} & \num{51.30} & \num{71.83} \\
		\midrule
		Linear & \num{63.07} & \num{61.17} & \num{45.22} & \num{60.26} & \num{76.33} \\
		\methodName$_\text{linear}$ & \underline{\num{66.33}} & \num{63.92} & \bf \num{52.15} & \num{62.35} & \underline{\num{78.36}} \\
		\midrule
		MLP & \num{63.00} & \num{64.77} & \num{49.01} & \underline{\num{62.61}} & \num{77.45} \\
		Transformer & \num{63.98} & \underline{\num{66.16}} & \num{47.32} & \num{61.50} & \num{76.86} \\
		\methodName$_\text{ft}$ & \bf \num{68.10} & \bf \num{66.79} & \underline{\num{51.89}} & \bf \num{64.83} & \bf \num{79.34} \\
		\midrule
		FT$_\text{reported}$ & \num[round-mode=places,round-precision=1]{67.2} & \num[round-mode=places,round-precision=1]{73.6} & \num[round-mode=places,round-precision=1]{73.3} & - & - \\
		\bottomrule
	\end{tabularx}
 }
 \caption{\textbf{HMDB Results.} Linear and non-linear probing accuracies on HMDB-51~\cite{kuehne2011hmdb}. Despite the small size of the dataset, we see that \methodName$_\text{ft}$ is outperforming all the other methods, and in the case of SVT, it even outperforms the fully fine-tuned model.
    We also see that \methodName$_\text{linear}$ outperforms the Transformer in most cases with only a single linear layer (the best linear accuracy in the literature is 66.7\%~\cite{recasens2021broaden}). Similar to UCF results, we see a considerable increase in the performance of $k$-NN classifier for pre-trained MAE-based models.}
    \label{table:hmdb}
\end{table}

\noindent\textbf{UCF-101 \& HMDB-51:} For these smaller datasets, besides linear and non-linear probing, we also use $k$-NN probing (see Table~\ref{table:ucf} and Table~\ref{table:hmdb}).
With \methodName$_{k\text{-NN}}$, we see a consistent improvement over the baseline and find that pre-trained MAE-based models greatly benefit from our training. 
This can be explained by the additional invariance properties introduced through the SET loss term in \methodName training.
Across the board, we also see that in the case of linear probing, not only does \methodName$_{linear}$ outperform Linear, but it also outperforms Transformer, which leverages many more parameters. 
In the case of SVT, our \methodName$_{linear}$ also outperforms the best reported linear accuracy on UCF101 (92.7\% vs. 92.6\%~\cite{recasens2021broaden}). Finally, \methodName$_{ft}$ achieves better results than all the non-linear baselines and even outperforms the fully fine-tuned SVT (68.1\% vs. 67.2\%).

\noindent\textbf{Kinetics-400:} We present our main results on Kinetics-400~\cite{kay2017kinetics} in Table~\ref{table:kinetics}. Our \methodName$_\text{linear}$ with SVT backbone beats the previous state of the art (71.8\% vs. 71.5\%~\cite{feichtenhofer2021large}) and \methodName$_\text{ft}$ can even improve the accuracy of VMAE$^\text{B}_\text{ft}$, which is a strong supervised model, from 81.5\% to 81.84\%.

\begin{table}[t]
	\centering\small
 \resizebox{\linewidth}{!}{%
    \begin{tabularx}{0.5\textwidth}{@{}lccccc@{}}
		\toprule 
		{} & SVT & $\rho$BYOL & VMAE$^\text{B}$ & VMAE$^\text{L}$ & VMAE$^\text{B}_\text{ft}$\\
		\midrule
		Linear & \num{71.71} & \num{68.82} & \num{43.50} & \num{60.73} & \num{81.52}\\
		\methodName$_\text{linear}$ & \num{71.78} & \num{68.38} & \num{43.96} & \num{60.66} & \num{81.44}\\
		\midrule
		MLP & \num{71.19} & \underline{\num{69.42}} & \underline{\num{45.48}} & \num{61.64} & \num{81.27}\\
		Transformer & \underline{\num{72.18}} & \num{69.28} & \num{44.85} & \underline{\num{62.15}} & \underline{\num{81.70}}\\
		\methodName$_\text{ft}$ & \bf{\num{72.38}} & \bf{\num{69.63}} & \bf{\num{46.15}} & \bf{\num{62.67}} & \bf{\num{81.84}}\\
		\bottomrule
	\end{tabularx}
 }
 \caption{\textbf{Kinetics-400 Results.} Linear and non-linear probing accuracies on Kinetics-400~\cite{kay2017kinetics} without any extra data and using RGB frames only. 
    While \methodName$_\text{linear}$ is on par with Linear, we observe clear improvements for non-linear probing in the case of  \methodName$_\text{ft}$. 
    Note that the best linear accuracy on this dataset (without any extra data) is 71.5~\cite{feichtenhofer2021large} and the best full fine-tuning accuracy is 86.7~\cite{Wei2022MaskedFP}. 
 }
    \label{table:kinetics}
\end{table}

\begin{table}[t]
	\centering\small
  \resizebox{\linewidth}{!}{%
	\begin{tabularx}{0.5\textwidth}{@{}lccccc@{}}
		\toprule 
		{} & SVT & $\rho$BYOL & VMAE$^\text{B}$ & VMAE$^\text{L}$ & VMAE$^\text{B}_\text{ft}$\\
		\midrule
		Linear & \underline{\num{66.43}} & \num{56.43} & \num{31.25} & \num{48.42} & \num{79.79} \\
		\methodName$_\text{linear}$ & \num{65.96} & \num{57.74} & \bf{\num{34.03}} & \underline{\num{49.21}} & \underline{\num{79.94}} \\
		\midrule
		MLP & \num{65.44} & \num{58.68} & \num{30.47} & \num{48.27} & \num{79.37} \\
		Transformer & \num{64.97} & \underline{\num{58.95}} & \num{29.89} & \num{48.27} & \num{79.47} \\
		\methodName$_\text{ft}$ & \bf{\num{67.01}} & \bf{\num{59.52}} & \underline{\num{33.92}} & \bf{\num{50.36}} & \bf{\num{80.47}} \\
		\bottomrule
	\end{tabularx}
 }
 \caption{\textbf{Kinetics-400 Low-shot Results.} Linear and non-linear probing accuracies on 10\% of Kinetics-400~\cite{kay2017kinetics}. \methodName is more robust to the size of the labeled dataset. \methodName$_\text{ft}$ does not overfit like the other non-linear probes (MLP and Transformer) and outperforms the baselines.}
    \label{table:kinetics_lowshot}
\end{table}

Following the evaluation setup of self-supervised image representations~\cite{chen2020simple,caron2021emerging,zhou2021ibot}, we also introduce low-shot K400 video classifications by sampling 10 percent of the videos (in a class-balanced way) and training the probes only on those.
We still test on the whole evaluation set of K400.
This low-shot setting is more aligned with the typical use-case of self-supervised models in which there is abundant unlabeled data for training via self-supervision and a small set of labeled data for fine-tuning. 
Results in Table~\ref{table:kinetics_lowshot} show that our method is particularly effective in this low-shot setting.
While most other non-linear probes overfit and perform worse than the linear probes, our \methodName$_\text{ft}$ does not overfit and clearly outperforms the baselines.

\subsection{Ablations}
\label{subsec:ablations}
In this section, we start from a baseline setup consisting of a two-layer transformer with a hidden size of 256, 20\% chance of masking clips, trained with a batch size of 512 for 200 epochs, and using two sets of 8 views for representation learning.
Using \methodName$_\text{ft}$, we explore different loss functions, masking ratios, number of layers, and finally, the number of views during training and testing.
All experiments are performed on UCF and HMDB.

\begin{table}[t]
    \centering\small
      \resizebox{\linewidth}{!}{%

    \begin{tabularx}{0.5\textwidth}{@{}c>{\centering\arraybackslash}X>{\centering\arraybackslash}X>{\centering\arraybackslash}X>{\centering\arraybackslash}X>{\centering\arraybackslash}X@{}}	
    \toprule
		{} & {} & \multicolumn{2}{c}{UCF-101} & \multicolumn{2}{c}{HMDB-51} \\
		SET & \mcm & SVT & $\rho$BYOL & SVT & $\rho$BYOL\\  \midrule
		\cmark & \xmark & 91.80 & 92.20 & 64.50 & 63.59 \\
		\xmark & \cmark & 92.01 & \bf{93.81} & 62.81 & 64.05 \\
		\cmark & \cmark & \bf{93.20} & 92.99 & \bf{64.57} & \bf{65.61} \\
		\bottomrule
	\end{tabularx}
 }
 \caption{\textbf{Loss Function.} \methodName$_\text{ft}$ accuracy with different loss function combinations (the masking ratio here is 20\%).
    We can see that having \mcm is always beneficial, and the SET loss is almost always helpful. We use both loss terms for our final model.}
    \label{table:loss}	
\end{table}

\begin{table}[t]
	\centering\small
   \resizebox{\linewidth}{!}{%

	\begin{tabularx}{0.5\textwidth}{@{}c>{\centering\arraybackslash}X>{\centering\arraybackslash}X>{\centering\arraybackslash}X>{\centering\arraybackslash}X}
		\toprule
		{} & \multicolumn{2}{c}{UCF-101} & \multicolumn{2}{c}{HMDB-51} \\
		Masking Ratio & SVT & $\rho$BYOL & SVT & $\rho$BYOL\\  \midrule
		0.15 & 93.18 & 93.25 & 64.37 & 64.83 \\ 
		0.25 & \bf{93.20} & \bf{93.81} & \bf{65.49} & \bf{65.62} \\ 
		0.35 & 93.15 & 93.06 & 64.18 & 65.22 \\ 
		0.45 & 92.96 & 93.02 & 63.39 & 64.96 \\ 
		\bottomrule
	\end{tabularx}
}
\caption{\textbf{Masking Ratio.} \methodName$_\text{ft}$ accuracy with different masking ratios. We observe best results around 25\% similar to NLP models~\cite{devlin2018bert} (15\%), and different from low-level video models like VideoMAE~\cite{tong2022videomae} (90\%).}
    \label{table:masking}
\end{table}

\noindent\textbf{Loss Function:} As explained in the method section, we have two loss terms, and each of them can be enabled or disabled for the pretraining. In Table~\ref{table:loss} we show that having both loss terms is better than the individual loss terms.

\noindent\textbf{Masking Ratio:} Masking ratio is an important hyperparameter and depends on the data modality, for example, BERT~\cite{devlin2018bert} uses 15\%, MSN~\cite{assran2022masked} uses 30\% (for ViT-Base), MAE~\cite{he2022masked} uses 75\%, and VideoMAE~\cite{tong2022videomae} uses 90 to 95\% masking. Since our clip representations are somewhat abstract representations of the video, we expect the optimal masking ratio to be close to NLP models rather than video MAEs. We have observed a steady decrease in the pretraining task's performance with higher masking ratios, so we only tested low masking ratios in Table~\ref{table:masking} and found out that 25\% is the optimal masking ratio.

\begin{table}[t]
	\centering\small
    \resizebox{0.97\linewidth}{!}{%

    \begin{tabularx}{0.5\textwidth}{@{}c>{\centering\arraybackslash}X>{\centering\arraybackslash}X>{\centering\arraybackslash}X>{\centering\arraybackslash}X>{\centering\arraybackslash}X@{}}
    \toprule
		Hidden & Num & \multicolumn{2}{c}{UCF-101} & \multicolumn{2}{c}{HMDB-51} \\
		Dim & Layers & SVT & $\rho$BYOL & SVT & $\rho$BYOL\\  \midrule
		64 & 1 & - & 92.62 & - & 63.16 \\
		128 & 1 & - & 92.83 & - & 63.68 \\
		256 & 1 & - & 92.86 & - & 64.39 \\
		128 & 2 & 92.78 & 93.52 & 63.26 & 65.55 \\
		256 & 2 & \bf{93.20} & \bf{93.81} & 65.49 & \bf{65.62} \\
		512 & 2 & 92.57 & 93.66 & \bf{65.68} & 64.77 \\
		128 & 3 & 92.33 & 93.25 & 64.83 & 65.55 \\
		256 & 3 & 92.75 & 93.52 & 65.49 & 65.16 \\
		512 & 3 & 92.86 & 92.93 & 65.49 & 64.84 \\
		\bottomrule
	\end{tabularx}
 }
 \caption{\textbf{Transformer Capacity.} \methodName$_\text{ft}$ accuracy with different model capacities. Having more than one transformer layer and not too few hidden channels is necessary for the best performance.}
    \label{table:transformer}	
\end{table}

\vspace{\baselineskip}

\noindent\textbf{Transformer Capacity:} We also explore the number of transformer layers and their hidden size in Table~\ref{table:transformer}. We can see that having more than one transformer layer is necessary for good results and too few hidden channels can hurt performance. 
However, there is a trade-off, and deeper transformers can lead to worse performance.

\begin{table}[t]
	\centering\small
    \resizebox{0.97\linewidth}{!}{%

	\begin{tabularx}{0.5\textwidth}{@{}c>{\centering\arraybackslash}X>{\centering\arraybackslash}X>{\centering\arraybackslash}X>{\centering\arraybackslash}X>{\centering\arraybackslash}X@{}}
		\toprule
		Num & Batch & \multicolumn{2}{c}{UCF-101} & \multicolumn{2}{c}{HMDB-51} \\
		Views & Size & SVT & $\rho$BYOL & SVT & $\rho$BYOL\\  \midrule
		4 $\times$ 2 & 256 & 92.65 & 92.83 & 64.35 & 64.24 \\
		6 $\times$ 2 & 256 & 92.70 & 93.07 & 64.57 & 64.37 \\
		8 $\times$ 2 & 256 & 92.80 & 93.49 & 64.64 & 64.63 \\
		4 $\times$ 2 & 512 & 92.67 & 93.49 & 64.85 & 64.50 \\
		6 $\times$ 2 & 512 & 93.18 & 93.68 & 64.90 & 65.35 \\
		8 $\times$ 2 & 512 & \bf{93.20} & \bf{93.81} & \bf{65.49} & \bf{65.62} \\
		4 $\times$ 2 & 1024 & 92.75 & 93.36 & 64.77 & 65.15 \\
		6 $\times$ 2 & 1024 & 92.96 & 93.57 & 64.96 & 65.48 \\
		8 $\times$ 2 & 1024 & 93.07 & OOM & 65.29 & OOM \\
		\bottomrule
	\end{tabularx}
 }
 \caption{\textbf{Number of Views.} \methodName$_\text{ft}$ accuracy with different numbers of clips and batch sizes. More views lead to consistent improvement, and large batch sizes are not necessary because of the hard negative samples.}
    \label{table:num_views}
\end{table}

\noindent\textbf{Number of Views:} Finally, we studied the model performance as we changed the number of views and batch size fed to the model. As can be seen in Table~\ref{table:num_views}, having more views has a large and consistent impact on the performance, and since we have hard negatives for contrastive loss within the video, we are not too reliant on large batch sizes.

\section{Conclusion}
\label{sec:conclusions}

In this paper, we introduced \methodName, a framework for video representation learning by aggregating the information from multiple clips at the same time. We combine contrastive learning and masked modeling with intuitions from predictive coding to obtain improved global and local representations of clips starting from frozen backbones. We evaluated these features using a wide array of backbones on different action classification and video understanding datasets and achieved strong or state-of-the-art results. The computational efficiency of our method is extremely useful for videos and opens the possibility to a wider group of researchers to work on video representation learning than previously possible. We also believe that working with a set of clips is an interesting direction for representation learning. Finally, as a surprising and maybe alarming observation, even contrastive representations that were trained to be invariant to data augmentations and spatio-temporal crops can be used for contrastive masked modeling. This might be due to benign memorization~\cite{Anagnostidis2022TheCC}, and understanding why this phenomenon happens might lead to a better understanding of contrastive learning.

\section{Acknowledgments}
This work was supported by grant 200020\_188690 of the Swiss National Science Foundation.

{\small
\bibliographystyle{ieee_fullname}
\bibliography{egbib}
}

\end{document}